\documentclass[entropy,article,accept,moreauthors,pdftex]{Definitions/mdpi}

\firstpage{1}
\pubvolume{21}
\issuenum{7}
\articlenumber{674}
\pubyear{2019}
\copyrightyear{2019}

\history{Received: 14 June 2019; Accepted: 8 July 2019; Published: 10 July 2019}

\usepackage[ruled]{algorithm2e}
\usepackage{wrapfig}
\graphicspath{{img/}}

\newcommand{\E}{\mathbb{E}}
\newcommand{\R}{\mathbb{R}}
\renewcommand{\P}{\mathcal{P}}
\DeclareMathOperator*{\argmax}{arg\,max}

\DeclareMathOperator*{\argsup}{arg\,sup}
\DeclareMathOperator{\dom}{dom}
\DeclareMathOperator{\range}{range}

\Title{Entropic Regularization of Markov Decision Processes}

\Author{Boris Belousov $^{1,*}$\href{https://orcid.org/0000-0001-7172-9104}{\orcidicon} and Jan Peters $^{1,2}$\href{https://orcid.org/0000-0002-5266-8091}{\orcidicon}}

\AuthorNames{Boris Belousov and Jan Peters}

\address{%
$^{1}$ \quad Department of Computer Science,
Technische Universität Darmstadt,
64289 Darmstadt, Germany\\
$^{2}$ \quad Max Planck Institute for Intelligent Systems, 72076 Tübingen, Germany}

\corres{Correspondence: belousov@ias.tu-darmstadt.de; Tel.: +49-6151-16-25387}

\abstract{
An optimal feedback controller for a given Markov decision process (MDP)
can in principle be synthesized by value or policy iteration.
However, if the system dynamics and the reward function are unknown,
a learning agent must discover an optimal controller via direct interaction with the environment.
Such interactive data gathering commonly leads to divergence towards dangerous
or uninformative regions of the state space unless additional regularization measures are taken.
Prior works proposed bounding the information loss
measured by the Kullback–Leibler (KL) divergence
at every policy improvement step to eliminate instability in the learning dynamics.
In this paper, we~consider a broader family of $f$-divergences,
and more concretely $\alpha$-divergences,
which inherit the beneficial property
of providing the policy improvement step in closed form
at the same time yielding a corresponding dual objective for policy evaluation.
Such entropic proximal policy optimization view gives a unified perspective on compatible actor-critic architectures.
In particular, common least-squares value function estimation
coupled with advantage-weighted maximum likelihood policy improvement
is shown to correspond to the Pearson $\chi^2$-divergence penalty.
Other actor-critic pairs arise for various choices of the penalty-generating function $f$.
On a concrete instantiation of our framework with the $\alpha$-divergence,
we carry out asymptotic analysis of the solutions for different values of~$\alpha$ and
demonstrate the effects of the divergence function choice
on common standard reinforcement learning problems.
}


\begin{document}

\section{Introduction}

Sequential decision-making problems under uncertainty are described by the mathematical framework
of Markov decision processes (MDPs)~\cite{Puterman1994}.
The core problem in MDPs is to find an optimal policy---a mapping from states to actions which
maximizes the expected cumulative reward collected by an agent over its lifetime.
In reinforcement learning (RL), the agent is additionally assumed to have no prior knowledge
about the environment dynamics and the reward function~\cite{sutton1998reinforcement}.
Therefore, direct policy optimization in the RL setting can be seen
as a form of stochastic black-box optimization: the agent proposes a query point in the form of a policy,
the environment evaluates this point by computing the expected return,
after that the agent updates the proposal and the process repeats~\cite{deisenroth2013survey}.
There are two conceptual steps in this scheme
known as policy evaluation and policy improvement~\cite{Bellman1957}.
Both steps require function approximation in high-dimensional and continuous state-action spaces
due to the curse of dimensionality~\cite{Bellman1957}.
Therefore, statistical learning approaches are employed to approximate the value function of a policy
and to perform policy improvement based on the data collected from the environment.

In contrast to traditional supervised learning, in reinforcement learning,
the data distribution changes with every policy update.
State-of-the-art
generalized policy iteration algorithms~\cite{Kakade2001, Peters2010, Schulman2015, schulman2017proximal}
are mindful of this covariate shift problem~\cite{shimodaira2000improving},
taking active measures to account for it.
To smoothen the learning dynamics, these algorithms limit the
information loss between successive policy updates
as measured by the KL divergence or approximations thereof~\cite{Neu2017}.
In the optimization literature, such approaches are categorized
as proximal (or trust region) algorithms~\cite{Parikh2014}.

The choice of the divergence function determines the geometry
of the information manifold~\cite{nielsen2018elementary}.
Recently, in particular in the area of implicit generative modeling~\cite{Goodfellow2014},
the choice of the divergence function was shown to have
a dramatic effect both on the optimization performance~\cite{bottou2017geometrical}
and the perceptual quality of the generated data
when various $f$-divergences were employed~\cite{Nowozin2016}.
In this paper, we carry over the idea of using generalized
entropic proximal mappings~\cite{Teboulle1992} given by an $f$-divergence to reinforcement learning.
We show that relative entropy policy search~\cite{Peters2010},
framed as an instance of stochastic mirror descent~\cite{nemirovski1983,Beck2003} as suggested by~\cite{Neu2017},
can be extended to use any divergence measure from the family of $f$-divergences.
The resulting algorithm provides insights into the compatibility
of policy and value function update rules in actor-critic architectures,
which we exemplify on several instantiations of the generic $f$-divergence
with representatives from the parametric family
of $\alpha$-divergences~\cite{chernoff1952measure,Amari1985,Cichocki2010}.

\section{Background}
This section provides the necessary background on policy gradients~\cite{deisenroth2013survey}
and entropic penalties~\cite{Teboulle1992} for later derivations and analysis.
Standard RL notation~\cite{thomas2015notation} is used throughout.

\subsection{Policy Gradient Methods}
Policy search algorithms~\cite{deisenroth2013survey}
commonly use the gradient estimator of the following form~\cite{Sutton1999}
\begin{equation}
\label{PG}
\hat{g} = \hat{E}_t \left[
\nabla_\theta \log \pi_\theta \hat{A}^w_t
\right]
\end{equation}
where $\pi_\theta(a|s)$ is a stochastic policy and $\hat{A}^w_t(s_t, a_t)$
is an estimator of the advantage function at timestep~$t$.
Expectation $\hat{E}_t[\dots]$ indicates an empirical average over a finite batch of samples,
in an algorithm that alternates between sampling and optimization.
The advantage estimate $\hat{A}^w_t$ in~\eqref{PG} can be obtained
from an estimate of the value function~\cite{Peters2008a, Schulman2016},
which in its turn is found by least-squares estimation.
Specifically, if $V^w(s)$ denotes a parametric value function,
and if $\hat{V}_t = \sum_{k = 0}^\infty \gamma^k R_{t+k}$ is taken as its rollout-based estimate,
then the parameters $w$ can be found as
\begin{equation}
\label{PE}
w = \arg\min_{\tilde{w}}\;
\hat{E}_t \left[ \| V^{\tilde{w}}(s_t) - \hat{V}_t \|^2 \right].
\end{equation}

The advantage estimate $\hat{A}_t^w = \sum_{k=0}^\infty \gamma^k \delta^w_{t+k}$
is then obtained by summing the temporal difference errors
$\delta^w_t = R_t + \gamma V^w(s_{t+1}) - V^w(s_t)$, also known as the Bellman residuals.
Treating $\hat{A}^w_t$ as fixed for the purpose of policy improvement,
we can view \eqref{PG} as the gradient of an advantage-weighted log-likelihood;
therefore, the policy parameters $\theta$ can be found as
\begin{equation}
\label{PI}
\theta = \arg\max_{\tilde{\theta}}\;
\hat{E}_t \left[
\log \pi_{\tilde{\theta}} \hat{A}^w_t
\right].
\end{equation}

Thus, actor-critic algorithms that use the gradient estimator~\eqref{PG} to update the policy
can be viewed as instances of the generalized policy iteration scheme,
alternating between policy evaluation~\eqref{PE} and policy improvement~\eqref{PI}.
In the following, we will see that the actor-critic pair~\eqref{PE} and \eqref{PI},
that combines least-squares value function fitting
with linear-in-the-advantage-weighted maximum likelihood policy improvement,
is just one representative from a family of such actor-critic pairs
arising for different choices of the $f$-divergence
penalty within our entropic proximal policy optimization framework.

\subsection{Entropic Penalties}
The term entropic penalties~\cite{Teboulle1992} refers to both $f$-divergences and Bregman divergences.
In this paper, we will focus on $f$-divergences, leaving generalization to Bregman divergences for future work.
The $f$-divergence~\cite{Csiszar1963} between two distributions $P$ and $Q$ with densities $p$ and $q$ is defined as
\begin{equation*}
D_f(p \| q) = E_q\left[ f\left(\frac{p}{q}\right) \right]
\end{equation*}
where $f$ is a convex function on $(0, \infty)$ with $f(1) = 0$
and $P$ is assumed to be absolutely continuous with respect to~$Q$.
For example, the KL divergence corresponds
to $f_1(x) = x \log x - (x - 1)$, with the formula also applicable to
unnormalized distributions~\cite{Zhu1995}.
Many common divergences lie on the curve of
$\alpha$-divergences~\cite{chernoff1952measure, Amari1985}
defined by a special choice of the generator function~\cite{Cichocki2010}
\begin{equation}
\label{alpha_div}
f_\alpha(x) = \frac{(x^\alpha-1) - \alpha(x-1)}{\alpha(\alpha-1)},
\quad \alpha \in \mathbb{R}.
\end{equation}

The $\alpha$-divergence $D_\alpha = D_{f_\alpha}$ will be used
as the primary example of the $f$-divergence throughout the paper.
For more details on the $\alpha$-divergence and its properties, see Appendix~\ref{app:background}.
Noteworthy is the symmetry of the $\alpha$-divergence with respect
to $\alpha = 0.5$, which relates reverse divergences as
$D_{0.5 + \beta}(p \| q) = D_{0.5 - \beta}(q \| p)$.

\section{Entropic Proximal Policy Optimization}
Consider the average-reward RL setting~\cite{sutton1998reinforcement},
where the dynamics of an ergodic MDP
are given by the transition density $p(s'|s,a)$.
An intelligent agent can modulate the system dynamics
by sampling actions~$a$~from a stochastic policy $\pi(a | s)$
at every time step of the evolution of the dynamical system.
The resulting modulated Markov chain with transition kernel
$p_\pi(s'|s) = \int_{A} p(s'|s,a) \pi(a|s) da$
converges to a stationary state distribution~$\mu_\pi(s)$
as time goes to infinity.
This stationary state distribution induces a state-action distribution
$\rho_\pi(s,a) = \mu_\pi(s) \pi(a|s)$, which corresponds to visitation
frequencies of state-action pairs~\cite{Puterman1994}.
The goal of the agent is to steer the system dynamics to desirable states.
Such objective is commonly encoded by the expectation of a random variable
$R \colon S \times A \to \mathbb{R}$ called reward in this context.
Thus, the agent seeks a policy that maximizes the expected reward
$J(\pi) = E_{\rho_\pi(s,a)}[R(s,a)]$.

In reinforcement learning, neither the reward function $R$ nor the
system dynamics $p(s'|s,a)$ are assumed to be known.
Therefore, to maximize (or even evaluate)
the objective $J(\pi)$, the agent must sample a batch of experiences
in the form of tuples $(s, a, r, s')$ from the dynamics
and use an empirical estimate
$\hat{J} = \hat{E}_t[R(s_t, a_t)]$
as a surrogate for the original objective.
Since the gradient of the expected reward with respect to the policy parameters
can be written as~\cite{Williams1992}
\begin{equation*}
\nabla_\theta J(\pi_\theta) =
E_{\rho_{\pi_\theta}(s,a)}
[\nabla_\theta \log \pi_\theta(a|s) R(s,a)]
\end{equation*}
with a corresponding sample-based counterpart
\begin{equation*}
\nabla_\theta \hat{J} =
\hat{E}_t [\nabla_\theta \log \pi_\theta(a_t|s_t) R(s_t,a_t)],
\end{equation*}
one may be tempted to optimize a sample-based objective
\begin{equation*}
\hat{E}_t [\log \pi_\theta(a_t|s_t) R(s_t,a_t)]
\end{equation*}
on a fixed batch of data $\{(s, a, r, s')_t\}_{t=1}^N$ till convergence.
However, such an approach ignores the fact that sampling distribution
$\rho_{\pi_\theta}(s,a)$ itself depends on the policy parameters $\theta$;
therefore, such greedy optimization
aims at a wrong objective~\cite{Peters2010}.
To have the correct objective, the dataset must be sampled anew
after every parameter update---doing otherwise will lead to overfitting and divergence.
This problem is known in statistics as the covariate shift problem~\cite{shimodaira2000improving}.

\subsection{Fighting Covariate Shift via Trust Regions}
A principled way to account for the change in the sampling distribution at every policy update step
is to construct an auxiliary local objective function that can be safely optimized till convergence.
Relative entropy policy search (REPS) algorithm~\cite{Peters2010}
proposes a candidate for such an objective
\begin{equation}
\label{REPS}
J_\eta(\pi) = E_{\rho_\pi}[R] -
\eta D_1(\rho_\pi \| \rho_{\pi_0})
\end{equation}
with $\pi_0$ being the current policy under which the data samples were collected,
policy $\pi$ being the improvement policy that needs to be found,
and $\eta > 0$ being a `temperature' parameter that determines
how much the next policy can deviate from the current one.
The original formulation employs a relative entropy trust region constraint $D_1$ with radius $\varepsilon$
instead of a penalty, which allows for finding the optimal temperature $\eta$
as a function of the trust region radius $\varepsilon$.

Importantly, the objective function~\eqref{REPS} can be optimized in closed
form for policy $\pi$ (i.e., treating the policy itself as a variable and not its parameters,
in contrast to standard policy gradients).
To that end, several constraints on $\rho_\pi$ are added
to ensure stationarity with respect to the given MDP~\cite{Peters2010}.
In a similar vein, we can solve Problem~\eqref{REPS} with respect to $\pi$
for any $f$-divergence with a twice differentiable generator function~$f$.

\subsection{Policy Optimization with Entropic Penalties}
Following the intuition of REPS,
we introduce an $f$-divergence penalized optimization problem
that the learning agent must solve at every policy iteration step
\begin{equation}
\label{EPPO}
\begin{aligned}
\underset{\pi}{\textrm{maximize}}
&\quad J_\eta(\pi) = E_{\rho_\pi}[R] -
\eta D_f(\rho_\pi \| \rho_{\pi_0}) \\
\textrm{subject to} &\quad \int_A \rho_\pi(s',a') da' =
\int_{S \times A} \rho_\pi(s,a) p(s'|s,a) ds da,
\quad \forall s' \in S, \\
&\quad \int_{S \times A} \rho_\pi(s,a) ds da = 1, \\
&\quad \rho_\pi(s,a) \geq 0, \quad \forall (s,a) \in S \times A.
\end{aligned}
\end{equation}
The agent seeks a policy that maximizes the expected reward
and does not deviate from the current policy too much.
The first constraint in \eqref{EPPO} ensures that the policy is compatible
with the system dynamics, and the latter two constraints ensure that $\pi$
is a proper probability distribution.
Please note that $\pi$ enters Problem~\eqref{EPPO} indirectly through $\rho_\pi$.
Since the objective has the form of free energy~\cite{Wainwright2007} in~$\rho_\pi$
with an $f$-divergence playing the role of the usual KL,
the solution can be expressed through the derivative
of the convex conjugate function $f_*^\prime$,
as shown for general nonlinear problems in~\cite{Teboulle1992},
\begin{equation}
\label{primal}
\rho_\pi(s, a) = \rho_{\pi_0}(s, a) f_*^\prime \left( \frac{
R(s,a) + \int_S V(s') p(s'|s,a) ds' - V(s) - \lambda + \kappa(s,a)
}{\eta}
\right).
\end{equation}
Here, $\{V(s), \lambda, \kappa(s,a)\}$ are the Lagrange dual variables
corresponding to the three constraints in~\eqref{EPPO}, respectively.
Although we get a closed-form solution for $\rho_\pi$, we still need
to solve the dual optimization problem to get the optimal dual variables
\begin{equation}
\label{dual}
\begin{aligned}
\underset{V, \lambda, \kappa}{\textrm{minimize}}
&\quad g(V, \lambda, \kappa) = \eta E_{\rho_{\pi_0}} \left[
f_*\left( \frac{A^V(s,a) - \lambda + \kappa(s,a)}{\eta} \right)
\right] + \lambda \\
\textrm{subject to} &\quad \kappa(s,a) \geq 0,
\quad \forall (s,a) \in S \times A, \\
&\quad \arg{f_*} \in \range_{x\,\geq\,0}{f^\prime(x)},
\quad \forall (s,a) \in S \times A.
\end{aligned}
\end{equation}
Remarkably, the advantage function $A^V(s,a) = R(s,a) + \int_S V(s') p(s'|s,a) ds' - V(s)$
emerges automatically in the dual objective.
The advantage function also appears in the penalty-free linear programming formulation
of policy improvement~\cite{Puterman1994},
which corresponds to the zero-temperature limit~$\eta \to 0$ of our formulation.
Thanks to the fact that the dual objective in~\eqref{dual} is given as an
expectation with respect to $\rho_{\pi_0}$, it can be straightforwardly estimated from rollouts.
The last constraint in~\eqref{dual} on the argument of $f_*$
is easy to evaluate for common $\alpha$-divergences.
Indeed, the convex conjugate~$f_\alpha^*$ of the generator function~\eqref{alpha_div} is given by
\begin{equation}
\label{alpha_div_dual}
f_{\alpha}^*(y)
=\frac{1}{\alpha}(1+(\alpha-1)y)^{\frac{\alpha}{\alpha-1}}
-\frac{1}{\alpha}, \quad \textrm{for} \;\; y(1-\alpha) < 1.
\end{equation}
Thus, the constraint on $\arg f_*$ in~\eqref{alpha_div}
is just a linear inequality $y(1-\alpha) < 1$ for any $\alpha$-divergence.

\subsection{Value Function Approximation}
For small grid-world problems, one can solve Problem~\eqref{dual} exactly
for $V(s)$. However, for larger problems or if the
state space is continuous, one must resort to function approximation.
Assume we plug an expressive function approximator $V^w(s)$
in~\eqref{dual}, then vector $w$ becomes a new vector of parameters
in the dual objective. Later, it will be shown that minimizing the dual
when $\eta \to \infty$ is closely related to minimizing the mean squared Bellman error.

\subsection{Sample-Based Algorithm for Dual Optimization}
To solve Problem~\eqref{dual} in practice, we gather a batch of samples
from policy $\pi_0$ and replace the expectation in the objective with
a sample average. Please note that in principle one also needs to estimate
the expectation of the future rewards $\int_S V(s') p(s'|s,a) ds'$.
However, since the probability of visiting the same state-action pair
in continuous space is zero, one commonly estimates this integral
from a single sample~\cite{deisenroth2013survey},
which is equivalent to assuming deterministic system dynamics.
Inequality constraints in~\eqref{dual} are linear and they must be
imposed for every $(s,a)$ pair in the dataset.

\subsection{Parametric Policy Fitting}
Assume Problem~\eqref{dual} is solved on a current batch of data sampled from $\pi_0$
and thus the optimal dual variables $\{V(s),\lambda, \kappa(s,a)\}$ are given.
Equation~\eqref{primal} allows one to evaluate the new density
$\rho_\pi(s,a)$ on any pair $(s, a)$ from the dataset.
However, it does not yield the new policy $\pi$ directly because
representation~\eqref{primal} is variational.
A common approach~\cite{deisenroth2013survey}
is to assume that the policy is represented by a parameterized conditional
density $\pi_\theta(a|s)$ and fit this density to the data using
maximum likelihood.

To fit a parametric density $\pi_\theta(a|s)$
to the true solution $\pi(a|s)$ given by~\eqref{primal},
we minimize the KL divergence $D_1(\rho_\pi \| \rho_{\pi_\theta})$.
Minimization of this KL is equivalent to maximization of the weighted maximum likelihood
$\hat{E}[f_*^\prime(\dots) \log \rho_{\pi_\theta}]$.
Unfortunately, distribution
$\rho_{\pi_\theta}(s,a) = \mu_{\pi_\theta}(s) \pi_\theta(a|s)$
is in general not known because $\mu_{\pi_\theta}(s)$
does not only depend on the policy but also on the system dynamics.
Assuming the effect of policy parameters on the stationary state distribution is small~\cite{deisenroth2013survey},
we arrive at the following optimization problem for fitting the policy parameters
\begin{equation}
\label{WML}
\theta = \arg\max_{\tilde{\theta}}\;
\hat{E}_t \left[
\log \pi_{\tilde{\theta}}(a_t|s_t) f_*^\prime \left(
\frac{
\hat{A}^w(s_t, a_t) - \lambda + \kappa(s_t, a_t)
}{\eta}
\right)
\right].
\end{equation}
Compare our policy improvement step~\eqref{WML} to the commonly used
advantage-weighted maximum likelihood (ML) objective~\eqref{PI}.
They look surprisingly similar
(especially if $f_*^\prime(y) = y$ is a linear function),
which is not a coincidence and will be systematically explained in the next sections.

\subsection{Temperature Scheduling}
The `temperature' parameter $\eta$ trades off reward vs divergence,
as can be seen in the objective function in Problem~\eqref{EPPO}.
In practice, devising a schedule for $\eta$ may be hard
because $\eta$ is sensitive to reward scaling and policy parameterization.
A more intuitive way to impose the $f$-divergence proximity condition
is by adding it as a constraint $D_f(\rho_\pi \| \rho_{\pi_0}) \leq \varepsilon$ with a fixed $\varepsilon$
and then treating the temperature $\eta \geq 0$ as an optimization variable.
Such formulation is easy to incorporate into the dual~\eqref{dual}
by adding a term $\eta\varepsilon$ to the objective and a constraint
$\eta \geq 0$ to the list of constraints.
Constraint-based formulation was successfully used before
with a KL divergence constraint~\cite{Peters2010}
and with its quadratic approximation~\cite{Kakade2001,Schulman2015}.

\subsection{Practical Algorithm for Continuous State-Action Spaces}
Our proposed approach for entropic proximal policy optimization
is summarized in Algorithm~\ref{algorithm}.
Following the generalized policy iteration scheme, we
(i) collect data under a given policy,
(ii) evaluate the policy by solving~\eqref{dual}, and
(iii) improve the policy by solving~\eqref{WML}.
In the following section,
several instantiations of Algorithm~\ref{algorithm}
with different choices of function~$f$ will be presented and studied.

\vspace{6pt}
\begin{algorithm}[H]
\SetAlgoLined
\KwIn{
Initial actor-critic parameters $(\theta_0, w_0)$,
divergence function $f$,
temperature $\eta > 0$}
\While{not converged}{
sample one-step transitions $\{(s, a, r, s')_t\}_{t=1}^N$
under current policy $\pi_{\theta_0}$\;
policy evaluation: optimize dual~\eqref{dual} with $V(s) = V^w(s)$
to obtain critic parameters $w$\;
policy improvement: perform weighted ML update~\eqref{WML}
to obtain actor parameters $\theta$\;}
\KwOut{
Optimal policy $\pi_\theta(a|s)$
and the corresponding value function $V^w(s)$}
\caption{
Primal-dual entropic proximal policy optimization
with function approximation\hspace*{-5mm}}
\label{algorithm}
\end{algorithm}

\section{High- and Low-Temperature Limits; $\alpha$-Divergences;
Analytic Solutions and Asymptotics}
How does the $f$-divergence penalty influence policy optimization?
How should one choose the generator function $f$?
What role does the step size play in optimization?
This section will try to answer these and related questions.
First, two special choices of the penalty function $f$ are presented,
which reveal that the common practice of
using mean squared Bellman error minimization
coupled with advantage reweighted policy update
is equivalent to imposing a Pearson $\chi^2$-divergence penalty.
Second, high- and low-temperature limits are studied,
on one hand revealing the special role the Pearson $\chi^2$-divergence plays,
being the high-temperature limit of all smooth $f$-divergences,
and on the other hand establishing a link to the linear programming formulation
of policy search as the low-temperature limit of our entropic penalty-based framework.

\subsection{KL Divergence ($\alpha = 1$)
and Pearson $\chi^2$-Divergence ($\alpha = 2$)}
As can be deduced from the form of~\eqref{WML}, great simplifications occur
when $f_*^\prime(y)$ is a linear function ($\alpha = 2$, see~\eqref{alpha_div_dual})
or an exponential function ($\alpha = 1$).
The fundamental reason for such simplifications lies in the fact that linear
and exponential functions are homomorphisms with respect to addition.
This allows, in particular, discovery of a closed-form solution
for the dual variable $\lambda$ and thus eliminate it from the optimization.
Moreover, in these two special cases, the dual variables $\kappa(s,a)$ can also be eliminated.
They are responsible for non-negativity of probabilities:
when $\alpha = 1$ (KL), $\kappa(s,a) = 0$ uniformly for all $\eta \geq 0$,
when $\alpha = 2$ (Pearson), $\kappa(s,a) = 0$ for sufficiently big $\eta$.
Table~\ref{table_alphas} gives the corresponding empirical actor-critic
optimization objective pairs.
A generic primal-dual actor-critic algorithm
with an $\alpha$-divergence penalty performs two steps
\begin{equation*}
\begin{array}{llrl}
(\textrm{step 1: policy evaluation}) &\quad&
\underset{w}{\textrm{minimize}} & \hat{g}_\alpha(w)\\
(\textrm{step 2: policy improvement}) &\quad&
\underset{\theta}{\textrm{maximize}} & \hat{L}_\alpha(\theta)
\end{array}
\end{equation*}
inside a policy iteration loop.
It is worth comparing the explicit formulas in Table~\ref{table_alphas}
to the customarily used objectives~\eqref{PE} and~\eqref{PI}.
To make the comparison fair, notice that~\eqref{PE} and~\eqref{PI}
correspond to discounted infinite horizon formulation with discount factor $\gamma \in (0, 1)$
whereas formulas in Table~\ref{table_alphas} are derived for the average-reward setting.
In general, the difference between these two settings can be ascribed
to an additional baseline that must be subtracted in the average
reward setting~\cite{sutton1998reinforcement}.
In our derivations, the baseline corresponds to the dual variable $\lambda$,
as in classical linear programming formulation of policy iteration~\cite{Puterman1994},
and it is automatically gets subtracted from the advantage (see~\eqref{dual}).
\begin{table}[H]
\caption{Empirical policy evaluation and policy improvement objectives
for $\alpha \in \{1, 2\}$.}
\centering
\begin{tabular}{ll}
\toprule
\textbf{KL Divergence ($\alpha = 1$)} &
\textbf{Pearson $\chi^2$-Divergence ($\alpha = 2$)}\\
\midrule
$
\hat{g}_1(w) = \eta \log \left( \hat{E}_t\left[
\exp\left( \frac{\hat{A}^w(s_t,a_t)}{\eta} \right)
\right] \right)
$&
$
\hat{g}_2(w) = \frac{1}{2\eta} \hat{E}_t \left[
\left(\hat{A}^w(s_t,a_t) - \hat{E}_t\left[ \hat{A}^w \right]\right)^2
\right]
$\\
$
\hat{L}_1(\theta) = \hat{E}_t \left[
\log \pi_{\theta}(a_t|s_t) \exp \left(
\frac{
\hat{A}^w(s_t, a_t) - \hat{g}_1(w)
}{\eta}
\right)
\right]
$&
$
\hat{L}_2(\theta) = \frac{1}{\eta} \hat{E}_t \left[
\log \pi_{\theta}(a_t|s_t) \left(
\hat{A}^w(s_t,a_t) - \hat{E}_t\left[ \hat{A}^w \right] + \eta
\right)
\right]
$\\
\bottomrule
\end{tabular}
\label{table_alphas}
\end{table}

\subsubsection*{Mean Squared Error Minimization with Advantage Reweighting is Equivalent to Pearson~Penalty}
The baseline for $\alpha = 2$ is given by the average advantage
$\lambda_2 = \hat{E}_t\left[ \hat{A}^w(s_t,a_t) \right]$,
which also equals the average return
in our setting~\cite{sutton1998reinforcement,Puterman1994}.
Therefore, to translate the formulas
from Table~\ref{table_alphas} to the
discounted infinite horizon form~\eqref{PE} and~\eqref{PI},
we need to remove the baseline
and add discounting to the advantage;
that is, set
$A^w(s,a) = R(s,a) + \gamma \int_S V^w(s') p(s'|s,a) ds' - V^w(s)$.
Then the dual objective
\begin{equation}
\label{PE_Pearson}
\hat{g}_2(w) \propto \hat{E}_t \left[
\left(\hat{A}^w(s_t,a_t) \right)^2
\right]
\end{equation}
is proportional to the average squared advantage.
Naive optimization of~\eqref{PE_Pearson} leads to the family of
residual gradient algorithms~\cite{Baird1995,Dann2014}.
However, if the same Monte Carlo estimate of the value function is used
as in~\eqref{PE}, then~\eqref{PE_Pearson} and~\eqref{PE} are exactly equivalent.
The same holds for the Pearson actor
\begin{equation}
\label{PI_Pearson}
\hat{L}_2(\theta) \propto \hat{E}_t \left[
\log \pi_{\theta}(a_t|s_t)
\hat{A}^w(s_t,a_t)
\right]
\end{equation}
and the standard policy improvement~\eqref{PI} provided that
$\eta = \hat{E}_t\left[ \hat{A}^w(s_t,a_t) \right]$.
That means~\eqref{PI_Pearson} is equivalent to~\eqref{PI} if
the weight of the divergence penalty is equal to the expected return.

\subsection{High- and Low-Temperature Limits}
In the previous subsection,
we established a direct correspondence
between the least-squares value function fitting
coupled with the advantage-weighted maximum likelihood
policy parameters estimation~\eqref{PE} and \eqref{PI}
and the dual-primal pair of optimization
problems~\eqref{PE_Pearson} and \eqref{PI_Pearson}
arising from our Algorithm~\ref{algorithm}
for the special choice of the Pearson $\chi^2$-divergence penalty.
In this subsection, we will show that this is not a coincidence
but a manifestation of the fundamental fact that
the Pearson~$\chi^2$-divergence is the quadratic approximation
of any smooth $f$-divergence about unity.

\subsubsection{High Temperatures: All Smooth $f$-Divergences Tend Towards Pearson $\chi^2$-Divergence}
There are two ways to show the independence of the primal-dual solution \eqref{dual}--\eqref{WML}
on the choice of the divergence penalty: either exactly solve an approximate problem
or approximate the exact solution of the original problem.
In the first case, the penalty is replaced with its Taylor expansion at $\eta\to\infty$,
which turns out to be the Pearson $\chi^2$-divergence, and then the derivation
becomes equivalent to the natural policy gradient derivation~\cite{Kakade2001}.
In the second case, the exact solution~\eqref{dual}--\eqref{WML} is expanded by Taylor:
for big $\eta$, dual variables~$\kappa(s,a)$ can be dropped if $\rho_{\pi_0}(s,a) > 0$, which yields
\begin{equation}
\label{conj_limit}
f_*\left( \frac{A^w(s,a) - \lambda}{\eta} \right) =
f_*(0) + \frac{A^w(s,a) - \lambda}{\eta} f_*^\prime(0)
+ \frac{1}{2} \left(\frac{A^w(s,a) - \lambda}{\eta}\right)^2
f_*^{\prime\prime}(0)
+ o\left(\frac{1}{\eta^2}\right).
\end{equation}
By definition of the $f$-divergence, the generator function~$f$
satisfies the condition $f(1) = 0$.
Without loss of generality~\cite{Sason2016},
one can impose an additional constraint $f^\prime(1) = 0$ for convenience.
Such constraint ensures that the graph of the function $f(x)$ lies
entirely in the upper half-plane, touching the $x$-axis at a single
point $x = 1$.
From the definition of the convex conjugate $f_*^\prime = (f^\prime)^{-1}$,
we can deduce that $f_*^\prime(0) = 1$ and $f_*(0) = 0$;
by rescaling, it is moreover possible to set
$f^{\prime\prime}(1) = f_*^{\prime\prime}(0) = 1$.
These properties are automatically satisfied by the $\alpha$-divergence,
which can be verified by a direct computation.
With this in mind, it is straightforward to see that
substitution of~\eqref{conj_limit} into~\eqref{dual}
yields precisely the quadratic objective $\hat{g}_2(w)$ from Table~\ref{table_alphas},
the difference being of the second order in $1/\eta$.

To obtain the asymptotic policy update objective,
one can expand~\eqref{WML} in the high-temperature limit $\eta \to \infty$
and observe that it equals $\hat{L}_2(\theta)$ from Table~\ref{table_alphas}
with the difference being of the second order in~$1/\eta$.
Therefore, it is established that the choice of the divergence function plays
a minor role for big temperatures (small policy update steps).
Since this is the mode in which the majority of iterative algorithms
operate, our entropic proximal policy optimization point of view
provides a rigorous justification for the common practice
of using the mean squared Bellman error objective for value function fitting
and the advantage-weighted maximum likelihood objective for policy improvement.

\subsubsection{Low Temperatures: Linear Programming Formulation Emerges in the Limit}
Setting $\eta$ to a small number is equivalent to allowing large
policy update steps because $\eta$ is the weight of the divergence
penalty in the objective function~\eqref{EPPO}.
Such regime is rather undesirable in reinforcement learning because
of the covariate shift problem mentioned in the introduction.
Problem~\eqref{EPPO} for $\eta \to 0$ turns into a well-studied
linear programming formulation~\cite{Puterman1994,Neu2017}
that can be readily applied if the model $\{p(s'|s,a), R(s,a)\}$ is known.

It is not straightforward to derive the asymptotics
of policy evaluation~\eqref{dual} and policy improvement~\eqref{WML}
for a general smooth $f$-divergence in the low-temperature limit $\eta \to 0$
because the dual variables $\kappa(s,a)$ do not disappear,
in contrast to the high-temperature limit~\eqref{conj_limit}.
However, for the KL divergence penalty (see Table~\ref{table_alphas}),
one can show that the policy evaluation objective $g_1(w)$
tends towards the supremum of the advantage $g_1(w) \to \sup_{s,a} A^w(s,a)$;
the optimal policy is deterministic,
$\pi(a|s) \to \delta(a - \arg \sup_{b} A^w(s,b))$,
therefore $L(\theta) \to \log \pi_\theta(\bar{a}|\bar{s})$
with $(\bar{s}, \bar{a}) = \arg \sup_{s',a'} A^w(s',a')$.

\section{Empirical Evaluations}
\label{sec:bandit-problem}
To develop an intuition regarding the influence of
the entropic penalties on policy improvement, we first consider
a simplified version of the reinforcement learning problem---namely
the stochastic multi-armed bandit problem~\citep{Bubeck2012}.
In this setting, our algorithm is closely related to the family of Exp3 algorithms~\citep{Auer2003},
originally motivated by the adversarial bandit problem.
Subsequently, we evaluate our approach in the standard reinforcement learning setting.

\subsection{Illustrative Experiments on Stochastic Multi-Armed Bandit Problems}
In the stochastic multi-armed bandit problem~\citep{Bubeck2012},
at every time step $t \in \{1,\dots,T\}$, an agent chooses among $K$ actions $a\in\mathcal{A}$.
After every choice $a_t = a$, it receives a noisy reward $R_t = R(a_t)$
drawn from a distribution with mean~$Q(a)$.
The goal of the agent is to maximize the expected total reward $J = E[\sum_{t=1}^T R_t]$.
Given the true values $Q(a)$, the optimal strategy is to always choose
the best action, $a^*_t = \argmax_a{Q(a)}$.
However, due to the lack of knowledge, the agent faces the exploration-exploitation dilemma.
A generic way to encode the exploration-exploitation trade-off
is by introducing a policy~$\pi_t$, i.e., a distribution from which the agent draws actions $a_t \sim \pi_t$.
Thus, the question becomes: given the current policy~$\pi_t$
and the current estimate of action values~$\hat{Q}_t$,
what should the policy~$\pi_{t+1}$ at the next
time step be? Unlike the choice of the best action under perfect information,
such sampling policies are hard to derive
from first principles~\citep{Ghavamzadeh2015}.

We apply our generic Algorithm~\ref{algorithm} to the stochastic multi-armed bandit problem
to illustrate the effects of the divergence choice.
The value function disappears because there is no state and no system dynamics in this problem.
Therefore, the estimate $\hat{Q}_t$ plays the role of the advantage,
and the dual optimization~\eqref{dual} is performed only with respect to the remaining Lagrange multipliers.

\subsubsection{Effects of $\alpha$ on Policy Improvement}
\begin{figure}[H]
\centering
\includegraphics[width=.98\textwidth]{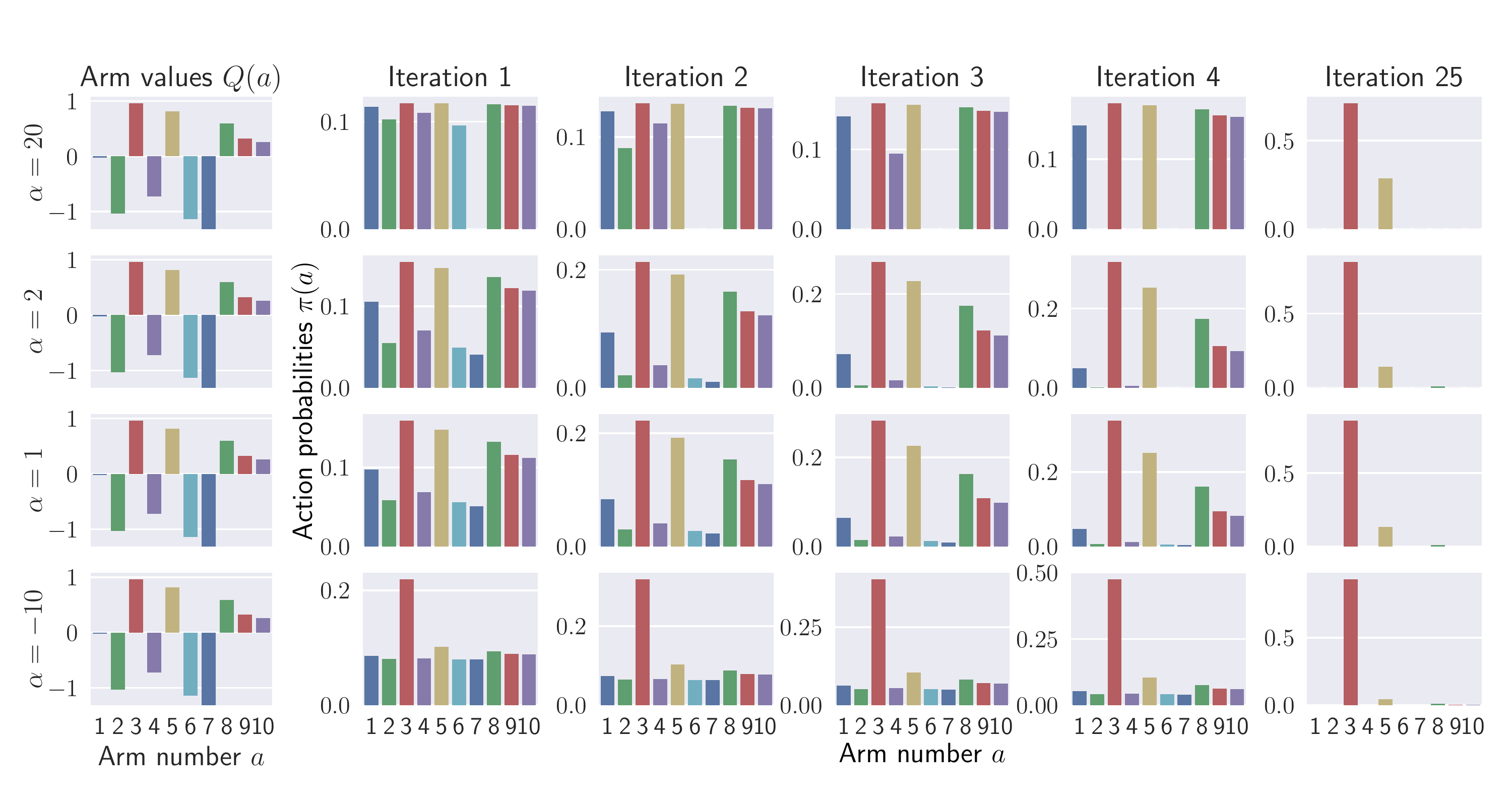}
\caption{
Effects of~$\alpha$ on policy improvement.
Each row corresponds to a fixed~$\alpha$.
First four iterations of policy improvement
together with a later iteration are shown in each row.
Large positive $\alpha$'s eliminate bad actions one by one,
keeping the exploration level equal among the rest.
Small $\alpha$'s weigh actions according to their values;
actions with low value get zero probability for~$\alpha > 1$,
but remain possible with small probability for~$\alpha \leq 1$.
Large negative $\alpha$'s focus on the best action,
exploring the remaining actions with equal probability.
}
\label{fig:bandit-alpha}
\end{figure}

Figure~\ref{fig:bandit-alpha} shows the effects of the $\alpha$-divergence choice on policy updates.
We consider a $10$-armed bandit problem with arm values $Q(a) \sim \mathcal{N}(0,1)$
and keep the temperature fixed at $\eta = 2$ for all values of $\alpha$.
Several iterations starting from an initial uniform policy are shown in the figure for comparison.
Extremely large positive and negative values of $\alpha$ result in $\varepsilon$-elimination and
$\varepsilon$-greedy policies, respectively.
Small values of $\alpha$, in contrast, weigh actions according to their values.
Policies for $\alpha < 1$ are peaked and heavy-tailed,
eventually turning into $\varepsilon$-greedy policies when $\alpha \to -\infty$.
Policies for $\alpha \geq 1$ are more uniform, but they put zero mass on bad actions,
eventually turning into $\varepsilon$-elimination policies when $\alpha \to \infty$.
For $\alpha \geq 1$, policy iteration may spend a lot of time in the end
deciding between two best actions, whereas for $\alpha < 1$ the final convergence is faster.

\subsubsection{Effects of $\alpha$ on Regret}

\begin{figure}[H]
\centering
\includegraphics[width=0.5\textwidth,trim={1.75em 0em 1.5em 0.5em},clip]{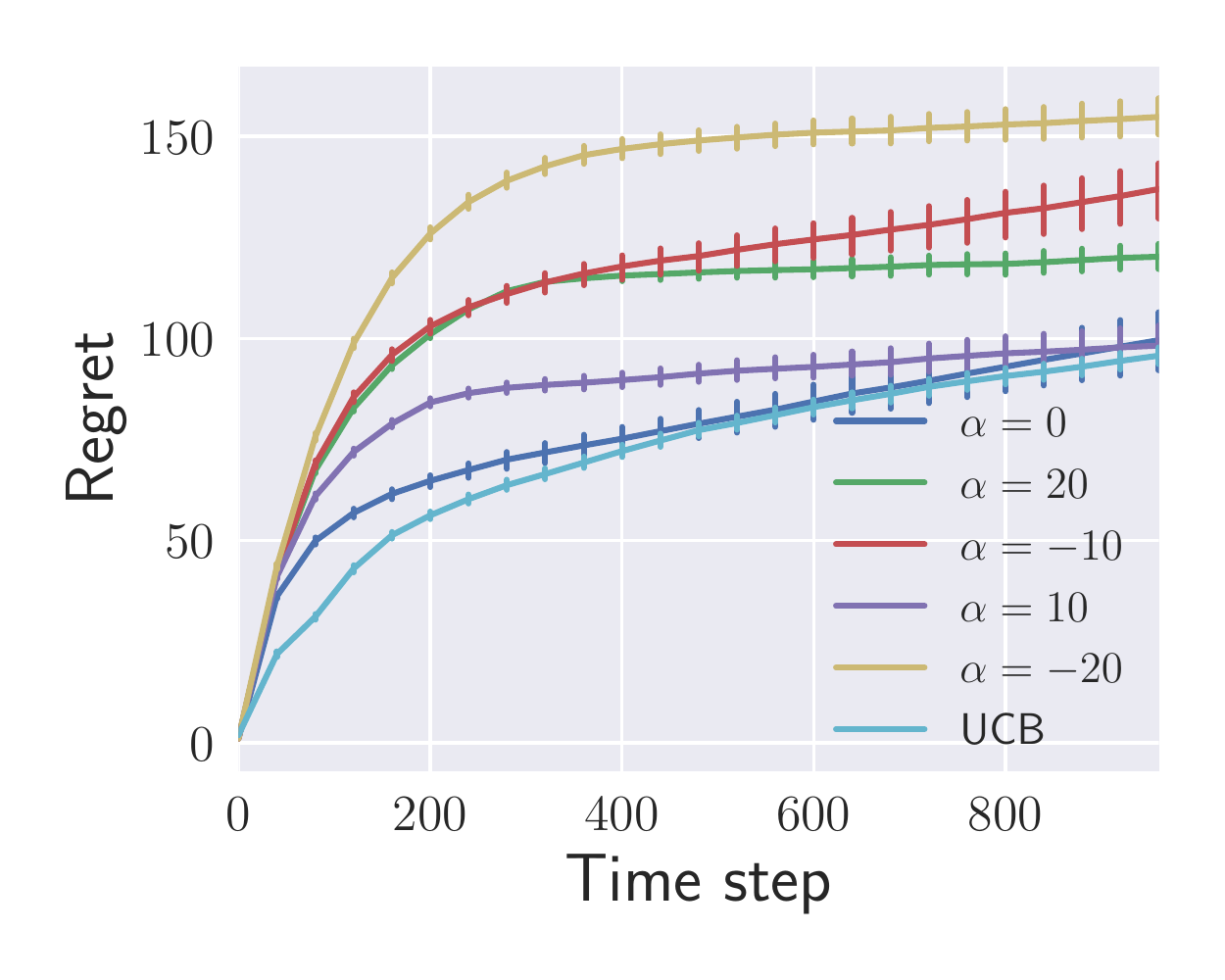}
\caption{Average regret for various values of $\alpha$.}
\label{fig:regret}
\end{figure}

The average regret $C_n = nQ_{\max} - \E [\sum_{t=0}^{n-1} R_t]$ is shown in Figure~\ref{fig:regret}
for different values of $\alpha$
as a function of the time step $n$ with $95\%$ confidence error bars.
The performance of the UCB algorithm~\citep{Bubeck2012} is also shown for comparison.
The presented results are obtained in a $20$-armed bandit environment
where rewards have Gaussian distribution $R(a) \sim \mathcal{N}(Q(a), 0.5)$.
Arm values are estimated from observed rewards and the policy
is updated every $20$ time steps.
The temperature parameter~$\eta$ is decreased starting from $\eta = 1$
after every policy update according to the schedule $\eta^+ = \beta \eta$
with $\beta=0.8$.
Results are averaged over $400$ runs.
In general, extreme $\alpha$'s accumulate more regret.
However, they eventually focus on a single action and flatten out.
Small $\alpha$'s accumulate less regret, but they may keep exploring
sub-optimal actions longer.
Values of $\alpha \in [0, 2]$ perform comparably with UCB
after around $400$ steps, once reliable estimates of values have been obtained.

\begin{figure}[H]
\centering
\includegraphics[width=0.5\textwidth,
trim={1.5em 0em 1em 1.25em},clip]{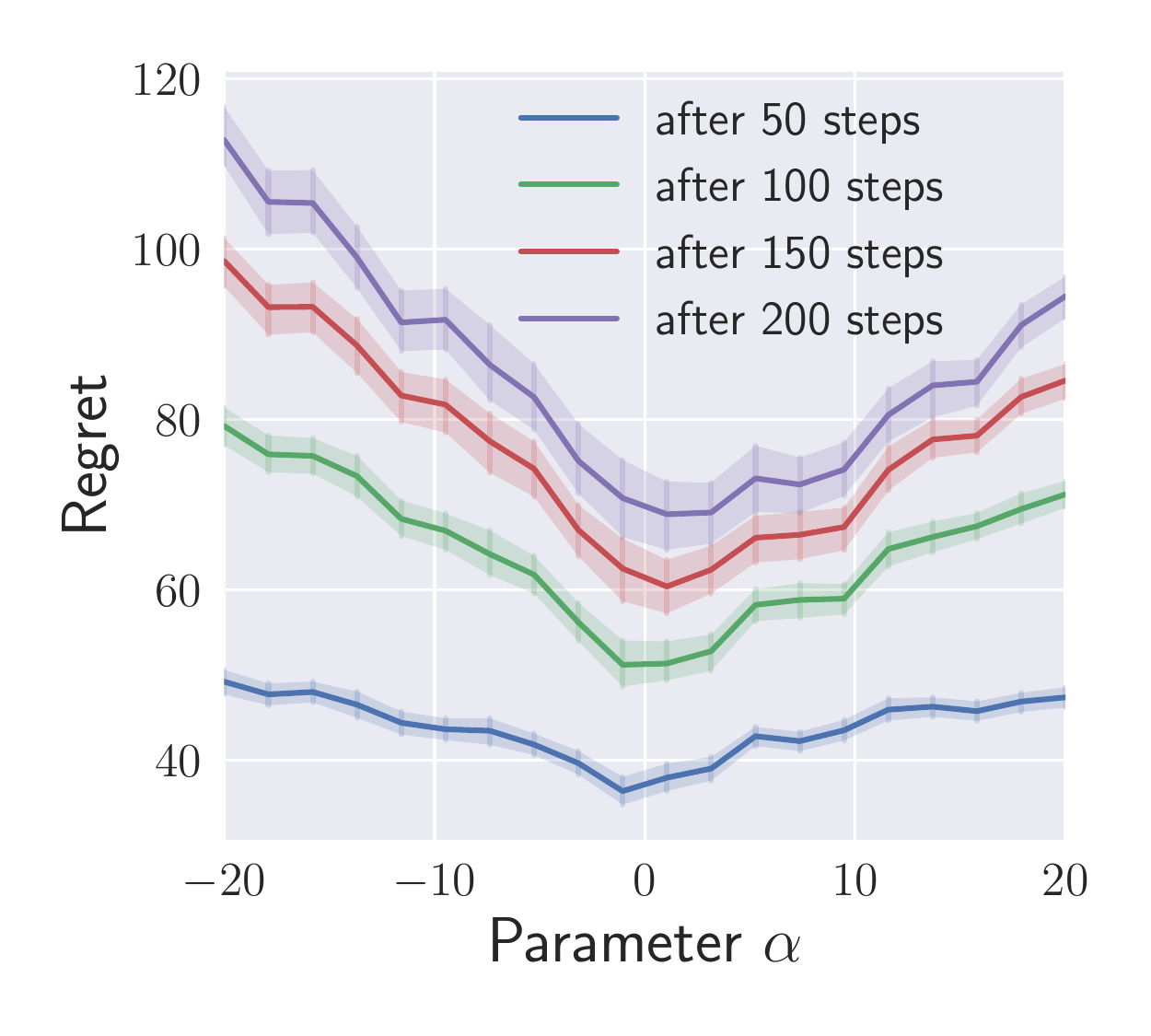}
\caption{Regret after a fixed time as a function of $\alpha$.}
\label{fig:regret_alpha}
\end{figure}

Figure~\ref{fig:regret_alpha} shows the average regret after a given number
of time steps as a function of the divergence type $\alpha$.
As can be seen from the figure, smaller values of~$\alpha$ result in lower regret.
Large negative $\alpha$'s correspond to $\varepsilon$-greedy policies,
which oftentimes prematurely converge to a sub-optimal action,
failing to discover the optimal action for a long time if the exploration
probability $\varepsilon$ is small.
Large positive $\alpha$'s correspond to $\varepsilon$-elimination policies,
which may by mistake completely eliminate the best action
or spend a lot of time deciding between two options in the end of learning,
accumulating more regret.
The optimal value of the parameter $\alpha$ depends on the time horizon
for which the policy is being optimized.
Depending on the horizon, the minimum of the curves shifts from slightly negative~$\alpha$'s
towards the range $\alpha \in [0, 2]$ with increasing time horizon.

\subsection{Empirical Evaluations on Ergodic MDPs}

\begin{figure}[H]
\centering
\includegraphics[width=.98\textwidth]{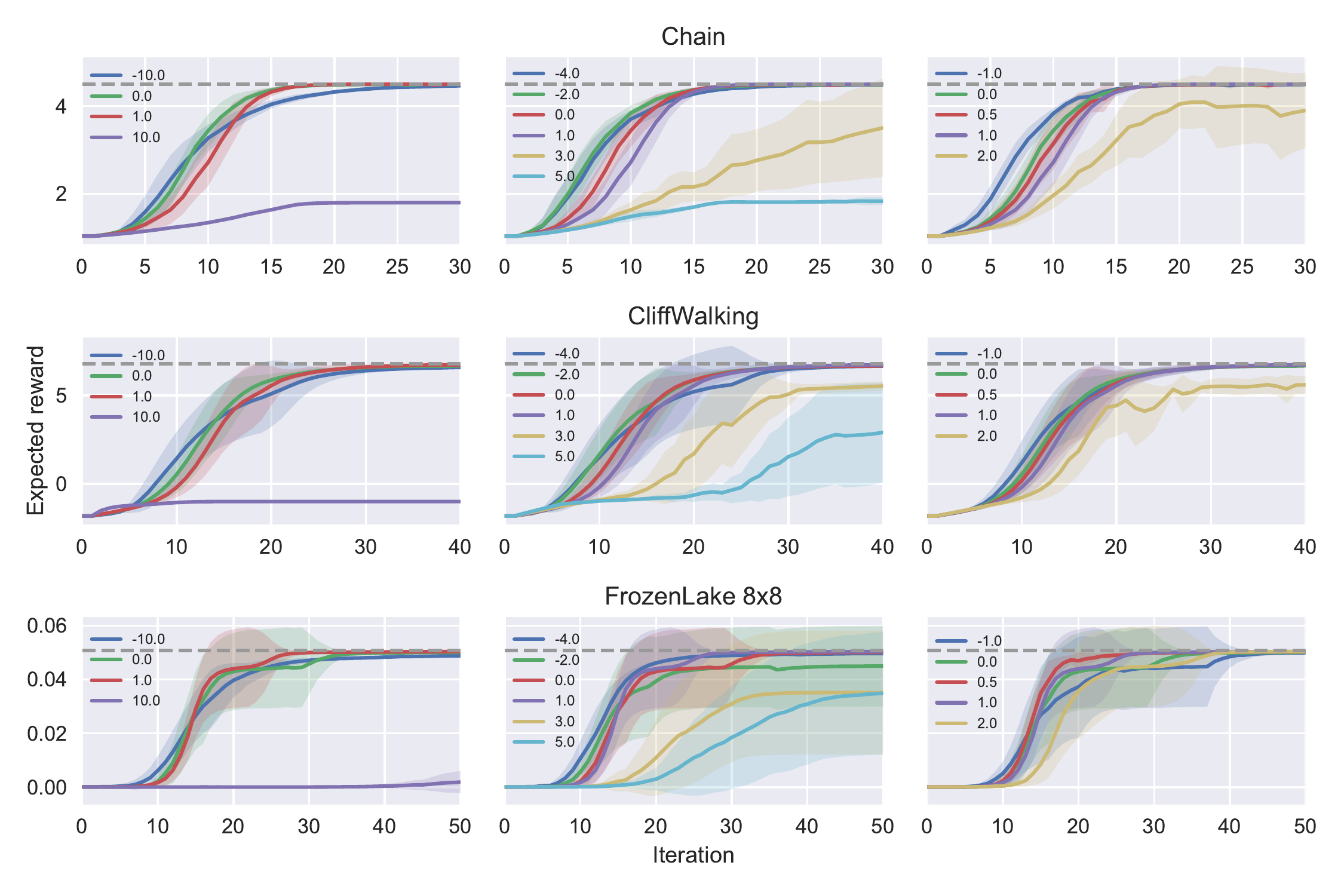}
\caption{
Effects of~$\alpha$-divergence on policy iteration.
Each row corresponds to a given environment.
Results for different values of~$\alpha$
are split into three subplots within each row,
from the more extreme~$\alpha$'s on the left
to the more refined values on the right.
In all cases, more negative values $\alpha < 0$ initially
show faster improvement because they immediately jump
to the mode and keep the exploration level low;
however, after a certain number of iterations they get overtaken
by moderate values $\alpha \in [0, 1]$ that weigh
advantage estimates more evenly.
Positive $\alpha > 1$ demonstrate high variance in the
learning dynamics because they clamp the probability
of good actions to zero if the advantage estimates
are overly pessimistic,
never being able to recover from such a mistake.
Large positive $\alpha$'s may even fail
to reach the optimum altogether,
as exemplified by $\alpha = 10$ in the plots.
The most stable and reliable $\alpha$-divergences
lie between the reverse KL ($\alpha=0$) and the KL ($\alpha=1$),
with the Hellinger distance ($\alpha=0.5$)
outperforming both on the FrozenLake environment.
}
\label{fig:RL-alpha}
\end{figure}

We evaluate our policy iteration algorithm with $f$-divergence on
standard grid-world reinforcement learning problems from OpenAI Gym~\citep{Brockman2016}.
The environments that terminate or have absorbing states are restarted
during data collection to ensure ergodicity.
Figure~\ref{fig:RL-alpha} demonstrates the learning dynamics
on different environments for various choices of the divergence function.
Parameter settings and other implementation details can be found in Appendix~\ref{app:sim_params}.
In summary, one can either promote risk averse behavior by choosing $\alpha < 0$,
which may, however, result in sub-optimal exploration,
or one can promote risk seeking behavior with $\alpha > 1$,
which may lead to overly aggressive elimination of options.
Our experiments suggest that the optimal balance
should be found in the range $\alpha \in [0, 1]$.
It should be noted that the effect of the $\alpha$-divergence on policy iteration
is not linear and not symmetric with respect to $\alpha=0.5$,
contrary to what one could have expected
given the symmetry of the $\alpha$-divergence as a function of $\alpha$.
For example, switching from $\alpha=-3$ to $\alpha=-2$ may have little
effect on policy iteration,
whereas switching from $\alpha=3$ to $\alpha=4$
may have a much more pronounced influence on the learning dynamics.

\section{Related Work}
Apart from computational advantages,
information-theoretic approaches provide a solid framework
for describing and studying aspects of intelligent behavior~\cite{tishby2011information},
from autonomy~\cite{bertschinger2008autonomy} and curiosity~\cite{still2012information}
to bounded rationality~\cite{genewein2015bounded} and game theory~\cite{wolpert2006information}.

Entropic proximal mappings were introduced in~\cite{Teboulle1992}
as a general framework for constructing approximation
and smoothing schemes for optimization problem.
Problem formulation~\eqref{EPPO} presented here can be considered
as an application of this general theory to policy optimization in Markov decision processes.
Following the recent work~\cite{Neu2017} that establishes links between popular in reinforcement learning
KL-divergence-regularized policy iteration algorithms~\cite{Peters2010,Schulman2015}
and a well-known in optimization stochastic mirror descent algorithm~\cite{nemirovski1983,Beck2003},
one can view our Algorithm~\ref{algorithm} as an analog of the mirror descent
with an $f$-divergence penalty.

Concurrent works~\cite{geist2019theory, li2019unified} consider similar regularized formulations,
although in the policy space instead of the state-action distribution space
and in the infinite horizon discounted setting instead of the average-reward setting.
The $\alpha$-divergence in its entropic form, i.e., when the base measure is a uniform distribution,
was used in several papers under the name
Tsallis entropy~\cite{nachum2018path, lee2019tsallis, lee2018sparse, lee2018maximum},
where its sparsifying effect was exploited in large discrete action spaces.

An alternative proximal reinforcement learning scheme was introduced in~\cite{mahadevan2014proximal}
based on the extragradient method for solving variational inequalities and leveraging operator splitting techniques.
Although the idea of exploiting proximal maps and updates in the primal and dual spaces is similar to ours,
regularization in~\cite{mahadevan2014proximal} is applied in the value function space
to smoothen generalized TD learning algorithms, whereas we study regularization in the primal space.

\section{Conclusions}
We presented a framework for deriving actor-critic algorithms
as pairs of primal-dual optimization problems resulting from
regularization of the standard expected return objective with
so-called entropic penalties in the form of an $f$-divergence.
Several examples with $\alpha$-divergence penalties
have been worked out in detail.
In the limit of small policy update steps, all $f$-divergences
with twice differentiable generator function $f$ are approximated
by the Pearson $\chi^2$-divergence, which was shown to yield the most
commonly used in reinforcement learning pair of actor-critic updates.
Thus, our framework provides a sound justification for the common practice
of minimizing mean squared Bellman error in the policy evaluation step
and fitting policy parameters by advantage-weighted maximum likelihood
in the policy improvement step.

In the future work, incorporating non-differentiable generator functions,
such as the absolute value that corresponds to the total variation distance,
may provide a principled explanation for the empirical success of the algorithms
not accounted for by our current smooth $f$-divergence framework, such as
the proximal policy optimization algorithm~\cite{schulman2017proximal}.
Establishing a tighter connection between online convex optimization that employs Bregman divergences
and reinforcement learning will likely yield both a deeper understanding of the optimization dynamics in RL
and allow for improved practical algorithms building on the firm fundament of optimization theory.

\vspace{6pt}


\authorcontributions{
Conceptualization, B.B. and J.P.; investigation, B.B. and J.P.; software, B.B.; supervision, J.P.; writing, B.B. and J.P.
}

\funding{
This project has received funding from the European Union’s Horizon 2020
research and innovation program under grant agreement No. 640554.
}

\acknowledgments{We thank Hany Abdulsamad for many insightful discussions.}

\conflictsofinterest{
The authors declare no conflict of interest.
}

\appendixtitles{no} 
\appendix
\section{}
\label{app:background}
This section provides the background on the~$f$-divergence,
the~$\alpha$-divergence, and the convex conjugate function,
highlighting the key properties required for our derivations.

The {$f$-divergence}~\citep{Csiszar1963,morimoto1963markov,Ali1966}
generalizes many similarity measures between
probability distributions~\citep{Sason2016}.
For two distributions~$\pi$~and~$q$ on a finite set $\mathcal{A}$,
the $f$-divergence is defined as
\begin{equation*}
D_f(\pi \| q) =
\sum_{a \in \mathcal{A}} q(a) f\left( \frac{\pi(a)}{q(a)} \right),
\end{equation*}
where $f$ is a convex function on $(0, \infty)$ such that $f(1) = 0$.
For example, the KL~divergence corresponds to $f_{KL}(x) = x \log x$.
Please note that $\pi$ must be absolutely continuous with respect
to~$q$ to avoid division by zero, i.e., $q(a) = 0$ implies
$\pi(a) = 0$ for all $a \in \mathcal{A}$.
We additionally assume $f$ to be continuously differentiable,
which includes all cases of interest for us.
The $f$-divergence can be generalized to \emph{unnormalized distributions}.
For example, the generalized KL divergence~\citep{Zhu1995}
corresponds to $f_1(x) = x \log x - (x - 1)$.
The derivations in this paper benefit from employing unnormalized distributions
and subsequently imposing the normalization condition as a constraint.

\begin{figure}[H]
\centering
\includegraphics[width=0.5\textwidth,
trim={4em 0em 2.75em 3em},clip]{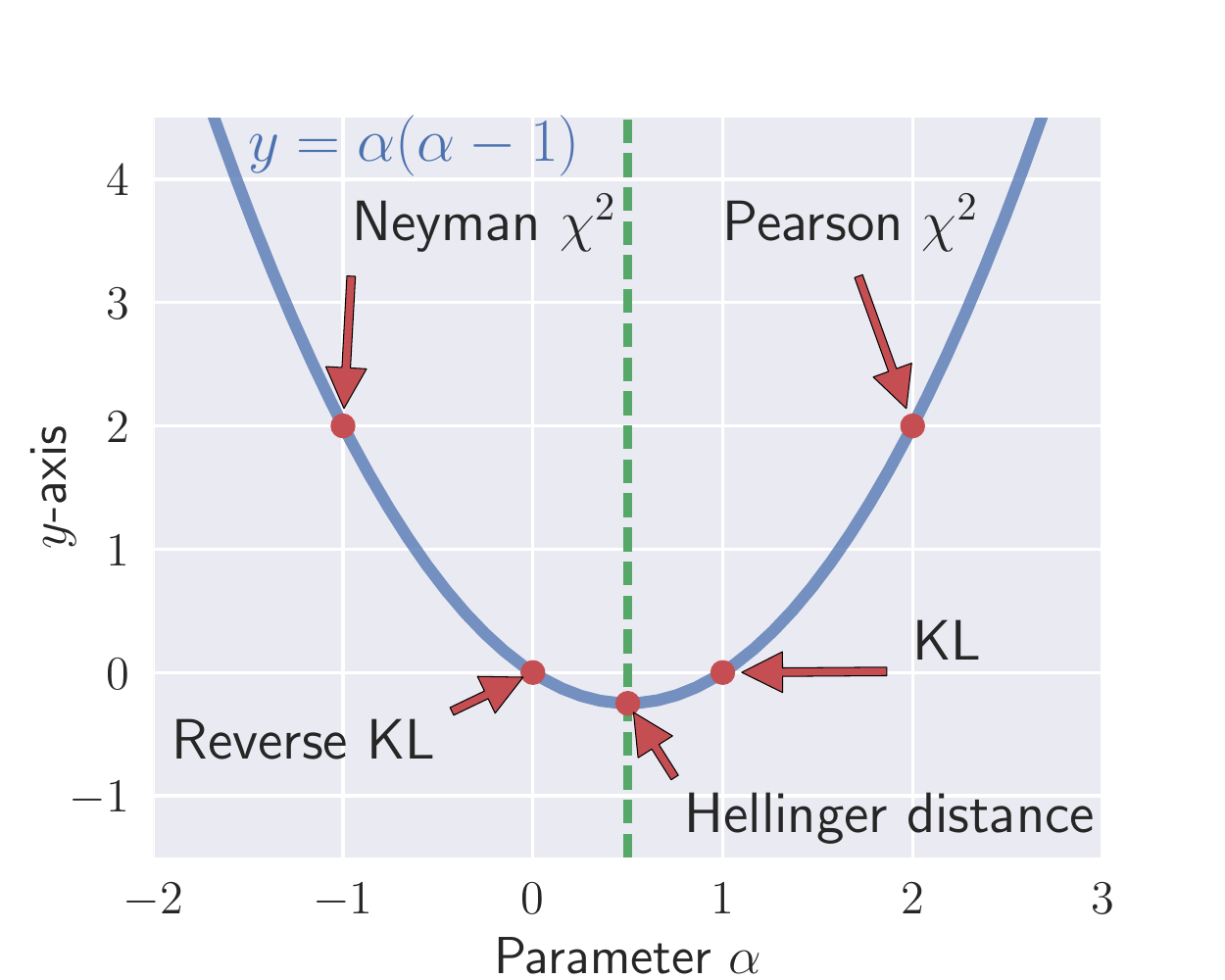}
\caption{\footnotesize
The $\alpha$-divergence smoothly connects several prominent divergences.}
\label{fig:alpha-divergence}
\end{figure}

The {$\alpha$-divergence}~\citep{chernoff1952measure, Amari1985}
is a one-parameter family of $f$-divergences
generated by the~\mbox{$\alpha$-function}
$f_\alpha(x)$ with  $\alpha \in \R$.
The particular choice of the family of functions $f_\alpha$ is motivated by
generalization of the natural logarithm~\citep{Cichocki2010}.
The $\alpha$-logarithm
$\log_\alpha(x) = (x^{\alpha - 1} - 1) / (\alpha - 1)$
is a power function for $\alpha \neq 1$ that turns into
the natural logarithm for $\alpha \to 1$.
Replacing the natural logarithm in the derivative
of the KL divergence $f_1^\prime = \log x$
by the $\alpha$-logarithm and integrating~$f_\alpha^\prime$
under the condition that~$f_\alpha(1) = 0$ yields the $\alpha$-function
\begin{equation}
\label{eq:alpha-function}
f_\alpha(x) = \frac{(x^\alpha - 1) - \alpha(x-1)}{\alpha(\alpha-1)}.
\end{equation}
The $\alpha$-divergence generalizes the KL divergence, reverse KL divergence,
Hellinger distance, Pearson~$\chi^2$-divergence,
and Neyman (reverse Pearson) $\chi^2$-divergence.
Figure~\ref{fig:alpha-divergence} displays
well-known $\alpha$-divergences as points
on the parabola $y = \alpha(\alpha-1)$.
For every divergence, there is a reverse
divergence symmetric with respect to the point $\alpha = 0.5$,
corresponding to the Hellinger distance.

The {convex conjugate} of $f(x)$ is defined as
$f^*(y) = \sup_{x \in \dom f} \{\langle y, x \rangle - f(x)\}$,
where the angle brackets $\langle y, x \rangle$
denote the dot product~\citep{Boyd2004}.
The key property $(f^*)^\prime = (f^\prime)^{-1}$
relating the derivatives of~$f^*$ and~$f$
yields Table~\ref{tab:divergences},
which lists common functions $f_\alpha$
together with their convex conjugates and derivatives.
In the general case~(\ref{eq:alpha-function}),
the convex conjugate and its derivative are given by
\begin{align}
f_{\alpha}^{*}(y)
&=\frac{1}{\alpha}(1+(\alpha-1)y)^{\frac{\alpha}{\alpha-1}}-
\frac{1}{\alpha}, \nonumber \\
(f_\alpha^*)^\prime(y)
&= \sqrt[\alpha-1]{1+\left(\alpha-1\right)y},
\quad\textrm{for}\;\;\textstyle y(1-\alpha)<1.
\label{eq:conjugate}
\end{align}
Function $f_\alpha$ is convex, non-negative,
and attains minimum at $x = 1$ with $f_\alpha(1) = 0$.
Function~$(f_\alpha^*)^\prime$ is positive on its domain
with $(f_\alpha^*)^\prime(0) = 1$.
Function~$f_\alpha^*$ has the property $f_\alpha^*(0) = 0$.
The linear inequality constraint (\ref{eq:conjugate})
on the $\dom f_\alpha^*$
follows from the requirement~$\dom f_\alpha = (0, \infty)$.
Another result from convex analysis crucial to our derivations is
Fenchel's equality
\begin{equation}
\label{eq:Fenchel}
f^*(y) + f(x^\star(y)) = \langle y, x^\star(y) \rangle,
\end{equation}
where $x^\star(y) = \argsup_{x \in \dom f} \{\langle y, x \rangle - f(x)\}$.
We will occasionally put the conjugation symbol at the bottom,
especially for the derivative of the conjugate
function $f_*^\prime = (f^*)^\prime$.

\begin{table}[t]
\tiny
\centering
\caption{\footnotesize Function $f_\alpha$,
its convex conjugate $f^*_\alpha$,
and their derivatives for some values of $\alpha$.
\label{tab:divergences}}
\begin{tabular*}{\textwidth}{l @{\extracolsep{\fill}} l l l l l l}
\toprule
\bf{Divergence} & \boldmath{$\alpha$} & \boldmath{$f(x)$} & \boldmath{$f^\prime(x)$} &
\boldmath{$(f^*)^\prime(y)$} & \boldmath{$f^*(y)$} & \boldmath{$\dom f^*$} \\
\midrule
KL & $1$ & $x\log x-(x-1)$ & $\log x$ & $e^{y}$ & $e^{y}-1$ & $\R$ \\
Reverse KL & $0$ & $-\log x+(x-1)$ & $-\frac{1}{x}+1$ &
$\frac{1}{1-y}$ & $-\log(1-y)$ & $y < 1$ \\
Pearson $\chi^2$ & $2$ & $\frac{1}{2}(x-1)^{2}$ & $x-1$ &
$y+1$ & $\frac{1}{2}(y+1)^{2}-\frac{1}{2}$ & $y > -1$ \\
Neyman $\chi^2$ & $-1$ & $\frac{(x-1)^{2}}{2x}$ &
$-\frac{1}{2x^{2}}+\frac{1}{2}$ &
$\frac{1}{\sqrt{1-2y}}$ & $-\sqrt{1-2y}+1$ & $y < \frac{1}{2}$ \\
Hellinger & $\frac{1}{2}$ & $2\left(\sqrt{x}-1\right)^{2}$ &
$2-\frac{2}{\sqrt{x}}$ & $\frac{4}{(2-y)^{2}}$ & $\frac{2y}{2-y}$ &
$y < 2$ \\
\bottomrule
\end{tabular*}
\vspace{-2em}
\end{table}

\section{}
\label{app:sim_params}
In all experiments, the temperature parameter $\eta$
is exponentially decayed $\eta_{i+1} = \eta_0 a^i$
in each iteration $i = 0, 1, \dots$.
The choice of $\eta_0$ and $a$ depends on the scale of the rewards
and the number of samples collected per policy update.
Tables for each environment list these parameters along
with the number of samples per policy update,
the number of policy iteration steps,
and the number of runs for averaging the results.
Where applicable, environment-specific settings are also listed. (see the Tables~\ref{table_chain}--\ref{table_frozen})
\begin{table}[H]
\caption{Chain environment.}
\centering
\begin{tabular}{ll}
\toprule
\textbf{Parameter}	& \textbf{Value}\\
\midrule
Number of states & 8 \\
Action success probability & 0.9 \\
Small and large rewards & (2.0, 10.0) \\
Number of runs & 10 \\
Number of iterations & 30 \\
Number of samples & 800 \\
Temperature parameters $(\eta_0, a)$ & (15.0, 0.9) \\
\bottomrule
\end{tabular}
\label{table_chain}
\end{table}
\unskip

\begin{table}[H]
\caption{CliffWalking environment.}
\centering
\begin{tabular}{ll}
\toprule
\textbf{Parameter}	& \textbf{Value}\\
\midrule
Punishment for falling from the cliff & $-10.0$ \\
Reward for reaching the goal & 100 \\
Number of runs & 10 \\
Number of iterations & 40 \\
Number of samples & 1500 \\
Temperature parameters $(\eta_0, a)$ & (50.0, 0.9) \\
\bottomrule
\end{tabular}
\label{table_cliff}
\end{table}
\unskip
\begin{table}[H]
\caption{FrozenLake environment.}
\centering
\begin{tabular}{ll}
\toprule
\textbf{Parameter}	& \textbf{Value}\\
\midrule
Action success probability & 0.8 \\
Number of runs & 10 \\
Number of iterations & 50 \\
Number of samples & 2000 \\
Temperature parameters $(\eta_0, a)$ & (1.0, 0.8) \\
\bottomrule
\end{tabular}
\label{table_frozen}
\end{table}

\reftitle{References}


\begin{thebibliography}{999}
\providecommand{\natexlab}[1]{#1}

\bibitem[Puterman(1994)]{Puterman1994}
Puterman, M.L.
\newblock {\em {Markov Decision Processes: Discrete Stochastic Dynamic
Programming}}; John Wiley \& Sons: Hoboken, NJ, USA, 1994. [\href{http://dx.doi.org/10.1080/00401706.1995.10484354}{CrossRef}]

\bibitem[Sutton and Barto(1998)]{sutton1998reinforcement}
Sutton, R.S.; Barto, A.G.
\newblock {\em {Reinforcement Learning: An Introduction}}; MIT Press: Cambridge, MA, USA,
1998.

\bibitem[Deisenroth {et~al.}(2013)Deisenroth, Neumann, and
Peters]{deisenroth2013survey}
Deisenroth, M.P.; Neumann, G.; Peters, J.
\newblock {A survey on policy search for robotics}.
\newblock {\em Found. Trends\textsuperscript{\textregistered} Robot.} {\bf 2013},
{\em 2},~1--142. [\href{http://dx.doi.org/10.1561/2300000021}{CrossRef}]

\bibitem[Bellman(1957)]{Bellman1957}
Bellman, R.
\newblock {Dynamic Programming}.
\newblock {\em{Science}} {\bf1957}, {\it 70}, 342. [\href{http://dx.doi.org/10.1108/eb059970}{CrossRef}]

\bibitem[Kakade(2001)]{Kakade2001}
Kakade, S.M.
\newblock {A Natural Policy Gradient}.
\newblock  {In Proceedings of the 14th International Conference on Neural Information Processing Systems: Natural and Synthetic}, Vancouver, BC, Canada, 3--8 December 2001;  pp.~1531--1538. [\href{http://dx.doi.org/10.1.1.19.8165}{CrossRef}]

\bibitem[Peters {et~al.}(2010)Peters, M{\"{u}}lling, and Altun]{Peters2010}
Peters, J.; M{\"{u}}lling, K.; Altun, Y.
\newblock {Relative Entropy Policy Search}.
\newblock  In Proceedings of the 24th AAAI Conference on Artificial Intelligence, Atlanta, GA, USA, 11--15 July 2010; pp. 1607--1612.

\bibitem[Schulman {et~al.}(2015)Schulman, Levine, Jordan, and
Abbeel]{Schulman2015}
Schulman, J.; Levine, S.; Moritz, P.; Jordan, M.; Abbeel, P.
\newblock {Trust Region Policy Optimization}.
\newblock  In Proceedings of the 32nd International Conference on International Conference on Machine Learning, Lille, France, 6--11~July~2015.

\bibitem[Schulman {et~al.}(2017)Schulman, Wolski, Dhariwal, Radford, and
Klimov]{schulman2017proximal}
Schulman, J.; Wolski, F.; Dhariwal, P.; Radford, A.; Klimov, O.
\newblock {Proximal policy optimization algorithms}.
\newblock {\em arXiv} {\bf 2017}, { arXiv:1707.06347}.

\bibitem[Shimodaira(2000)]{shimodaira2000improving}
Shimodaira, H.
\newblock {Improving predictive inference under covariate shift by weighting
the log-likelihood function}.
\newblock {\em J. Stat. Plann. Inference}. {\bf 2000},~227--244. [\href{http://dx.doi.org/10.1016/S0378-3758(00)00115-4}{CrossRef}]

\bibitem[Neu {et~al.}(2017)Neu, Jonsson, and G{\'{o}}mez]{Neu2017}
Neu, G.; Jonsson, A.; G{\'{o}}mez, V.
\newblock {A unified view of entropy-regularized Markov decision processes}.
\newblock  {\em arXiv} {\bf 2017}, {arXiv:1705.07798}.

\bibitem[Parikh(2014)]{Parikh2014}
Parikh, N.
\newblock {Proximal Algorithms}.
\newblock {\em Found. Trends\textsuperscript{\textregistered} Optim.} {\bf
2014}, {\em 1},~127--239. [\href{http://dx.doi.org/10.1561/2400000003}{CrossRef}]

\bibitem[Nielsen(2018)]{nielsen2018elementary}
Nielsen, F.
\newblock An elementary introduction to information geometry.
\newblock {\em arXiv} {\bf 2018}, {arXiv:1808.08271}.

\bibitem[Goodfellow \em{et~al.}(2014)Goodfellow, Pouget-Abadie, Mirza, Xu,
Warde-Farley, Ozair, Courville, and Bengio]{Goodfellow2014}
Goodfellow, I.; Pouget-Abadie, J.; Mirza, M.; Xu, B.; Warde-Farley, D.; Ozair,
S.; Courville, A.; Bengio, Y.
\newblock {Generative Adversarial Nets}.
\newblock  In Proceedings of the 27th International Conference on Neural Information Processing Systems, Montreal, QC, Canada, 8--13 December 2014.

\bibitem[Bottou \em{et~al.}(2017)Bottou, Arjovsky, Lopez-Paz, and
Oquab]{bottou2017geometrical}
Bottou, L.; Arjovsky, M.; Lopez-Paz, D.; Oquab, M.
\newblock {Geometrical Insights for Implicit Generative Modeling}.
\newblock {\em Braverman Read. Mach. Learn}. {\bf 2018}, {\em 11100}, 229--268.

\bibitem[Nowozin \em{et~al.}(2016)Nowozin, Cseke, and Tomioka]{Nowozin2016}
Nowozin, S.; Cseke, B.; Tomioka, R.
\newblock {f-GAN: Training Generative Neural Samplers using Variational
Divergence Minimization}.
\newblock  In Proceedings of the 30th International Conference on Neural Information Processing Systems, Barcelona, Spain, 5--10 December 2016; pp. 271--279.

\bibitem[Teboulle(1992)]{Teboulle1992}
Teboulle, M.
\newblock {Entropic Proximal Mappings with Applications to Nonlinear
Programming}.
\newblock {\em Math. Operations Res.} {\bf 1992}, {\em
17},~670--690. [\href{http://dx.doi.org/10.1287/moor.17.3.670}{CrossRef}]

\bibitem[Nemirovski and Yudin(1983)]{nemirovski1983}
Nemirovski, A.; Yudin, D.
\newblock {Problem complexity and method efficiency in optimization}. {\em J. Operational Res. Soc.} {\bf 1984}, {\em 35}, 455.

\bibitem[Beck and Teboulle(2003)]{Beck2003}
Beck, A.; Teboulle, M.
\newblock {Mirror descent and nonlinear projected subgradient methods for
convex optimization}.
\newblock {\em Operations Res. Lett.} {\bf 2003}, {\em 31},~167--175. [\href{http://dx.doi.org/10.1016/S0167-6377(02)00231-6}{CrossRef}]

\bibitem[Chernoff(1952)]{chernoff1952measure}
Chernoff, H.
\newblock {A measure of asymptotic efficiency for tests of a hypothesis based
on the sum of observations}.
\newblock {\em Ann. Math. Stat.} {\bf 1952}, {\em 23}, 493--507. [\href{http://dx.doi.org/10.1214/aoms/1177729330}{CrossRef}]

\bibitem[Amari(1985)]{Amari1985}
Amari, S.
\newblock {\em {Differential-Geometrical Methods in Statistics}}; Springer: New York, NY, USA, 1985. [\href{http://dx.doi.org/10.1007/978-1-4612-5056-2}{CrossRef}]

\bibitem[Cichocki and Amari(2010)]{Cichocki2010}
Cichocki, A.; Amari, S.
\newblock {Families of alpha- beta- and gamma- divergences: Flexible and robust
measures of Similarities}.
\newblock {\em Entropy} {\bf 2010}, {\em 12},~1532--1568. [\href{http://dx.doi.org/10.3390/e12061532}{CrossRef}]

\bibitem[Thomas and Okal(2015)]{thomas2015notation}
Thomas, P.S.; Okal, B.
\newblock {A notation for Markov decision processes}.
\newblock {\em arXiv} {\bf 2015}, arXiv:1512.09075.

\bibitem[Sutton \em{et~al.}(1999)Sutton, Mcallester, Singh, and
Mansour]{Sutton1999}
Sutton, R.S.; Mcallester, D.; Singh, S.; Mansour, Y.
\newblock {Policy Gradient Methods for Reinforcement Learning with Function
Approximation}.
\newblock In Proceedings of the 12th International Conference on Neural Information Processing Systems, Denver, CO, USA, 29 November--4 December 1999; pp. 1057--1063. [\href{http://dx.doi.org/10.1.1.37.9714}{CrossRef}]

\bibitem[Peters and Schaal(2008)]{Peters2008a}
Peters, J.; Schaal, S.
\newblock {Natural Actor-Critic}.
\newblock {\em Neurocomputing} {\bf 2008}, {\em 71},~1180--1190. [\href{http://dx.doi.org/10.1016/j.neucom.2007.11.026}{CrossRef}]

\bibitem[Schulman \em{et~al.}(2016)Schulman, Moritz, Levine, Jordan, and
Abbeel]{Schulman2016}
Schulman, J.; Moritz, P.; Levine, S.; Jordan, M.I.; Abbeel, P.
\newblock {High Dimensional Continuous Control Using Generalized Advantage
Estimation}.
\newblock  {\em arXiv} {\bf 2015}, arXiv:1506.02438.

\bibitem[Csisz{\'{a}}r(1963)]{Csiszar1963}
Csisz{\'{a}}r, I.
\newblock {Eine informationstheoretische Ungleichung und ihre Anwendung auf den
Beweis der Ergodizit{\"{a}}t von Markoffschen Ketten}.
\newblock {\em Publ. Math. Inst. Hungar. Acad. Sci.} {\bf 1963}, {\em
8},~85--108.

\bibitem[Zhu and Rohwer(1995)]{Zhu1995}
Zhu, H.; Rohwer, R.
\newblock {Information Geometric Measurements of Generalisation};
\newblock Technical Report; Aston University: Birmingham, UK, 1995.

\bibitem[Williams(1992)]{Williams1992}
Williams, R.J.
\newblock {Simple statistical gradient-following methods for connectionist
reinforcement learning}.
\newblock {\em Mach. Learn.} {\bf 1992}, {\em 8},~229--256. [\href{http://dx.doi.org/10.1007/BF00992696}{CrossRef}]

\bibitem[Wainwright and Jordan(2007)]{Wainwright2007}
Wainwright, M.J.; Jordan, M.I.
\newblock {Graphical Models, Exponential Families, and Variational Inference}.
\newblock {\em Found. Trends Mach. Learn.} {\bf 2007}, {\em
1},~1--305. [\href{http://dx.doi.org/10.1561/2200000001}{CrossRef}]

\bibitem[Baird(1995)]{Baird1995}
Baird, L.
\newblock {Residual Algorithms: Reinforcement Learning with Function
Approximation}.
\newblock {In Proceedings of the 12th International Conference on Machine
Learning}, Tahoe City, CA, USA, 9--12 July 1995; pp. 30--37. [\href{http://dx.doi.org/10.1.1.48.3256}{CrossRef}]

\bibitem[Dann \em{et~al.}(2014)Dann, Neumann, and Peters]{Dann2014}
Dann, C.; Neumann, G.; Peters, J.
\newblock {Policy Evaluation with Temporal Differences: A Survey and
Comparison}.
\newblock {\em J. Mach. Learn. Res.} {\bf 2014}, {\em
15},~809--883.

\bibitem[Sason and Verdu(2016)]{Sason2016}
Sason, I.; Verdu, S.
\newblock {F-divergence inequalities}.
\newblock {\em IEEE Trans. Inf. Theory} {\bf 2016}, {\em
62},~5973--6006. [\href{http://dx.doi.org/10.1109/TIT.2016.2603151}{CrossRef}]

\bibitem[Bubeck and Cesa-Bianchi(2012)]{Bubeck2012}
Bubeck, S.; Cesa-Bianchi, N.
\newblock {Regret Analysis of Stochastic and Nonstochastic Multi-armed Bandit
Problems}.
\newblock {\em Found. Trends Mach. Learn.} {\bf 2012}, {\em
5},~1--122. [\href{http://dx.doi.org/10.1561/2200000024}{CrossRef}]

\bibitem[Auer \em{et~al.}(2003)Auer, Cesa-Bianchi, Freund, and
Schapire]{Auer2003}
Auer, P.; Cesa-Bianchi, N.; Freund, Y.; Schapire, R.
\newblock {The Non-Stochastic Multi-Armed Bandit Problem}.
\newblock {\em SIAM J. Comput.} {\bf 2003}, {\em 32},~48--77. [\href{http://dx.doi.org/10.1137/S0097539701398375}{CrossRef}]

\bibitem[Ghavamzadeh \em{et~al.}(2015)Ghavamzadeh, Mannor, Pineau, and
Tamar]{Ghavamzadeh2015}
Ghavamzadeh, M.; Mannor, S.; Pineau, J.; Tamar, A.
\newblock {Bayesian Reinforcement Learning: A Survey}.
\newblock {\em Found. Trends Mach. Learn.} {\bf 2015}, {\em
8},~359--483. [\href{http://dx.doi.org/10.1561/2200000049}{CrossRef}]

\bibitem[Brockman \em{et~al.}(2016)Brockman, Cheung, Pettersson, Schneider,
Schulman, Tang, and Zaremba]{Brockman2016}
Brockman, G.; Cheung, V.; Pettersson, L.; Schneider, J.; Schulman, J.; Tang,
J.; Zaremba, W.
\newblock {OpenAI Gym}.
\newblock {\em arXiv} {\bf 2016}, {arXiv:1606.01540}.

\bibitem[Tishby and Polani(2011)]{tishby2011information}
Tishby, N.; Polani, D.
\newblock Information theory of decisions and actions. In {\em
Perception-Action Cycle}; Cutsuridis, V., Hussain, A., Taylor, J., Eds.; Springer: New York, NY, USA, 2011; pp. 601--636.

\bibitem[Bertschinger \em{et~al.}(2008)Bertschinger, Olbrich, Ay, and
Jost]{bertschinger2008autonomy}
Bertschinger, N.; Olbrich, E.; Ay, N.; Jost, J.
\newblock Autonomy: An information theoretic perspective.
\newblock {\em Biosystems} {\bf 2008}, {\em 91},~331--345. [\href{http://dx.doi.org/10.1016/j.biosystems.2007.05.018}{CrossRef}] [\href{http://www.ncbi.nlm.nih.gov/pubmed/17897774}{PubMed}]

\bibitem[Still and Precup(2012)]{still2012information}
Still, S.; Precup, D.
\newblock An information-theoretic approach to curiosity-driven reinforcement
learning.
\newblock {\em Theory Biosci.} {\bf 2012}, {\em 131},~139--148. [\href{http://dx.doi.org/10.1007/s12064-011-0142-z}{CrossRef}]

\bibitem[Genewein \em{et~al.}(2015)Genewein, Leibfried, Grau-Moya, and
Braun]{genewein2015bounded}
Genewein, T.; Leibfried, F.; Grau-Moya, J.; Braun, D.A.
\newblock Bounded rationality, abstraction, and hierarchical decision-making:
An information-theoretic optimality principle.
\newblock {\em Front. Rob. AI} {\bf 2015}, {\em 2},~27. [\href{http://dx.doi.org/10.3389/frobt.2015.00027}{CrossRef}]

\bibitem[Wolpert(2006)]{wolpert2006information}
Wolpert, D.H.
\newblock Information theory―the bridge connecting bounded rational game
theory and statistical physics. In {\em Complex Engineered Systems}; Braha, D., Minai, A., Bar-Yam, Y., Eds.; Springer: Berlin, Germany, 2006; pp. 262--290.

\bibitem[Geist {et~al.}(2019)Geist, Scherrer, and Pietquin]{geist2019theory}
Geist, M.; Scherrer, B.; Pietquin, O.
\newblock A Theory of Regularized Markov Decision Processes.
\newblock {\em arXiv} {\bf 2019}, {arXiv:1901.11275}.

\bibitem[Li {et~al.}(2019)Li, Yang, and Zhang]{li2019unified}
Li, X.; Yang, W.; Zhang, Z.
\newblock A Unified Framework for Regularized Reinforcement Learning.
\newblock {\em arXiv} {\bf 2019}, arXiv:1903.00725.

\bibitem[Nachum {et~al.}(2018)Nachum, Chow, and Ghavamzadeh]{nachum2018path}
Nachum, O.; Chow, Y.; Ghavamzadeh, M.
\newblock Path consistency learning in Tsallis entropy regularized MDPs.
\newblock {\em arXiv} {\bf 2018}, {arXiv:1802.03501}.


\bibitem[Lee {et~al.}(2019)Lee, Kim, Lim, Choi, and Oh]{lee2019tsallis}
Lee, K.; Kim, S.; Lim, S.; Choi, S.; Oh, S.
\newblock Tsallis Reinforcement Learning: A Unified Framework for Maximum
Entropy Reinforcement Learning.
\newblock {\em arXiv} {\bf 2019}, {arXiv:1902.00137}.

\bibitem[Lee \em{et~al.}(2018{\natexlab{a}})Lee, Choi, and Oh]{lee2018sparse}
Lee, K.; Choi, S.; Oh, S.
\newblock Sparse Markov decision processes with causal sparse Tsallis entropy
regularization for reinforcement learning.
\newblock {\em IEEE Rob. Autom. Lett.} {\bf 2018}, {\em
3},~1466--1473. [\href{http://dx.doi.org/10.1109/LRA.2018.2800085}{CrossRef}]

\bibitem[Lee \em{et~al.}(2018{\natexlab{b}})Lee, Choi, and Oh]{lee2018maximum}
Lee, K.; Choi, S.; Oh, S.
\newblock Maximum Causal Tsallis Entropy Imitation Learning.
\newblock  In Proceedings of the 32nd International Conference on Neural Information Processing Systems, Montreal, QC, Canada,  3--8 December 2018; pp. 4408--4418.

\bibitem[Mahadevan \em{et~al.}(2014)Mahadevan, Liu, Thomas, Dabney, Giguere,
Jacek, Gemp, and Liu]{mahadevan2014proximal}
Mahadevan, S.; Liu, B.; Thomas, P.; Dabney, W.; Giguere, S.; Jacek, N.; Gemp,
I.; Liu, J.
\newblock Proximal reinforcement learning: A new theory of sequential decision
making in primal-dual spaces.
\newblock {\em arXiv} {\bf 2014}, {arXiv:1405.6757}.

\bibitem[Morimoto(1963)]{morimoto1963markov}
Morimoto, T.
\newblock {Markov processes and the H-theorem}.
\newblock {\em J. Phys. Soc. Jpn.} {\bf 1963}, {\em
18},~328--331. [\href{http://dx.doi.org/10.1143/JPSJ.18.328}{CrossRef}]

\bibitem[Ali and Silvey(1966)]{Ali1966}
Ali, S.M.; Silvey, S.D.
\newblock {A General Class of Coefficients of Divergence of One Distribution
from Another}.
\newblock {\em J. R. Stat. Soc. Ser. B
(Methodol.)} {\bf 1966}, {\em 28},~131--142. [\href{http://dx.doi.org/10.1111/j.2517-6161.1966.tb00626.x}{CrossRef}]

\bibitem[Boyd and Vandenberghe(2004)]{Boyd2004}
Boyd, S.; Vandenberghe, L.
\newblock {\em {Convex Optimization}}; Cambridge University Press: Cambridge, UK, 2004; 487p. [\href{http://dx.doi.org/10.1017/CBO9780511804441}{CrossRef}]

\end{thebibliography}

\end{document}